\documentclass{article} 
\usepackage{iclr2020_conference,times}

\usepackage{amsmath,amsfonts,bm}

\def\eqref#1{equation~\ref{#1}}

\def\1{\bm{1}}

\DeclareMathAlphabet{\mathsfit}{\encodingdefault}{\sfdefault}{m}{sl}
\SetMathAlphabet{\mathsfit}{bold}{\encodingdefault}{\sfdefault}{bx}{n}

\usepackage{hyperref}
\usepackage{url}
\usepackage{graphicx}
\usepackage{float}
\title{Afro-MNIST: Synthetic generation of MNIST-style datasets for low-resource languages}

\iclrfinalcopy

\author{Daniel J. Wu \\
Department of Computer Science\\
Stanford University\\
\texttt{danjwu@stanford.edu} \\
\And
Andrew C. Yang \\
Department of Linguistics\\
Stanford University\\
\texttt{ycm@stanford.edu} \\
\And
Vinay Prabhu \\
UnifyID Inc. \\
\texttt{vinayup@gmail.com}
}

\iclrfinalcopy 
\begin{document}

\maketitle

\begin{abstract}
We present Afro-MNIST, a set of synthetic MNIST-style datasets for four orthographies used in Afro-Asiatic and Niger-Congo languages: Ge`ez (Ethiopic), Vai, Osmanya, and N'Ko.
These datasets serve as ``drop-in'' replacements for MNIST. We also describe and open-source a method for synthetic MNIST-style dataset generation from single examples of each digit. These datasets can be found at \url{https://github.com/Daniel-Wu/AfroMNIST}.
We hope that MNIST-style datasets will be developed for other numeral systems, and that these datasets vitalize machine learning education in underrepresented nations in the research community.
\end{abstract}

\begin{figure}[h]
    \hspace*{-1.25cm}
    \centering
    \includegraphics[width=\linewidth]{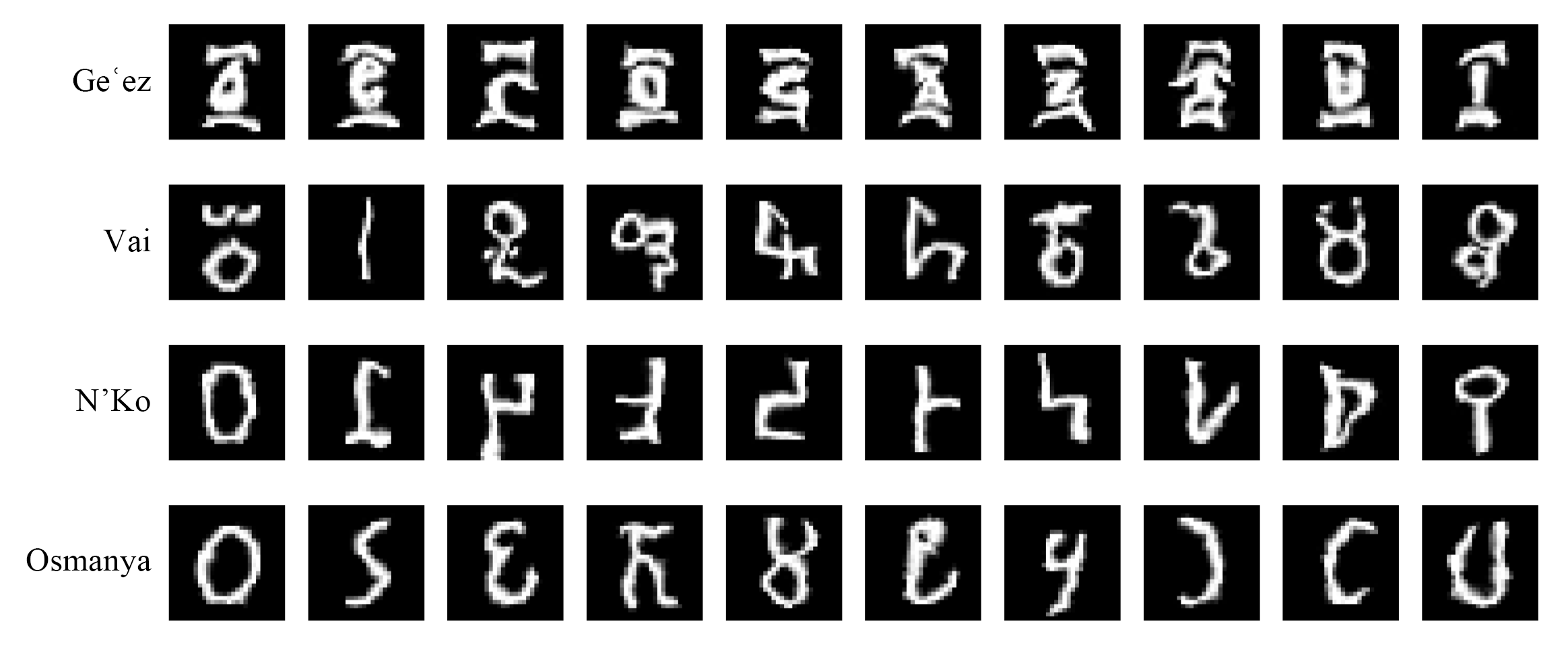}
    \caption{Afro-MNIST: Four synthetic MNIST-style datasets.}
    \label{fig:pullfig}
\end{figure}

\section{Motivation}
Classifying MNIST Hindu-Arabic numerals \citep{lecun1998gradient} has become the ``Hello World!'' challenge of the machine learning community. This task has excited a large number of prospective machine learning scientists and has led to practical advancements in optical character recognition. 

The Hindu-Arabic numeral system is the predominant numeral system used in the world today. However, there are a sizeable number of languages whose numeric glyphs are not inherited from the Hindu-Arabic numeral system. As it relates to human language, work in machine learning and artificial intelligence focuses almost exclusively on high-resource languages such as English and Mandarin Chinese.

Unfortunately, these ``mainstream'' languages comprise only a tiny fraction of all extant languages. Indeed, of over 7,000 languages in the world, the vast majority are not represented in the machine learning research community. In particular, we note that there is a vast collection of alternative numeral systems for which an MNIST-style dataset is not available. 

In an effort to make machine learning education more accessible to diverse groups of people, it is imperative that we develop datasets which represent the heterogeneity of existing numeral systems. Furthermore, as studied in \cite{mgqwashu2011academic}, familiarity of the glyph shapes from the students' mother tongue facilitates learning and enhances epistemological access. With the recent drive towards spreading AI literacy in Africa, we argue that it would be rather useful to use local numeral glyphs in the initial \textit{Machine Learning 101} courses to spark enthusiasm as well as to establish familiarity.

Additionally, we are inspired by the dire warning that appears in the work by \cite{mgqwashu2011academic} that \textit{the numeral system is the most endangered aspect of any language} and that the excuse of not having these datasets and downstream OCR applications will further accelerate the decline of several numeral systems dealt with in this work.

In similar work, \cite{prabhu2019kannada} collected and open-sourced an MNIST-style dataset for the Kannada language; however, this process took a significant amount of effort from 65 volunteers. While this methodology is effective for high-resource languages, we expect that it is not practical enough to be applied to the large number of low-resource languages. 

Much of the world's linguistic diversity comes from languages spoken in developing nations. In particular, there is a wealth of linguistic diversity in the languages of Africa, many of which have dedicated orthographies and numeral systems. In this work, we focus on the Ge`ez (Ethiopic), Vai, Osmanya, and N'Ko scripts\footnote{We note that the Vai, Osmanya, and N'Ko scripts are not in wide use, but nonetheless they can be synthesized using the methods we present.}, each of which have dedicated numerals. In particular, the Ge`ez script is used to transcribe languages such as Amharic and Tigrinya, which are spoken by some 30 million people \citep{eberhard_simons_fennig_2019}.

Because large amounts of training data for African languages such as Amharic and Somali are not readily available, we experiment with creating synthetic numerals that mimic the likeness of handwritten numerals in their writing systems. Previous work in few-shot learning and representation learning has shown that effective neural networks can be trained on highly perturbed versions of just a single image of each class \citep{exemplarnets}. Thus, we propose the synthetic generation of MNIST-style datasets from Unicode exemplars of each numeral.

\paragraph{Our Contributions:}
\begin{enumerate}
    \item We release synthetic MNIST-style datasets for four scripts used to write Afro-Asiatic or Niger-Congo languages: Ge`ez, Vai, Osmanya, N'Ko.
    \item We describe a general framework for resource-light syntheses of MNIST-style datasets.
\end{enumerate}

\section{Methodology}

\begin{figure}[h]
    \hspace*{-1.25cm}
    \centering
    \includegraphics[width=\linewidth]{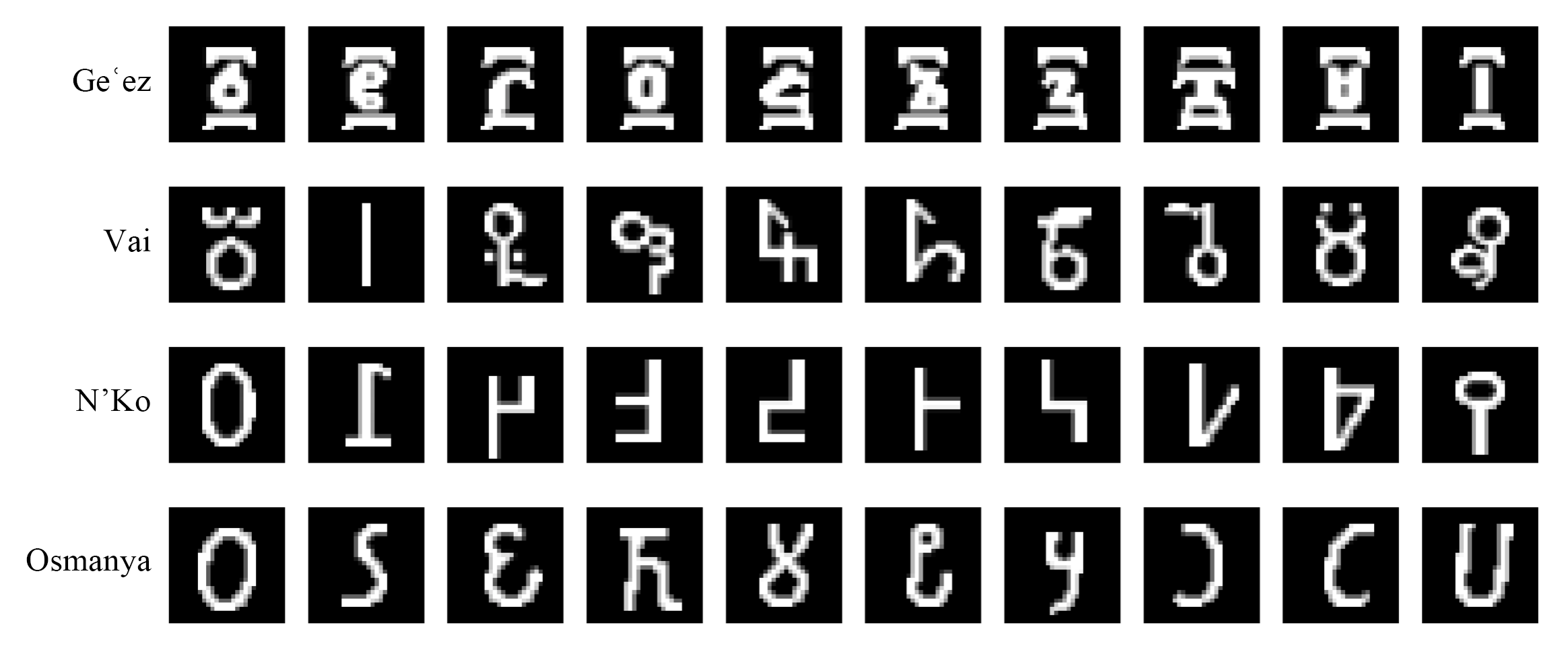}
    \caption{Exemplar images generated from Unicode characters}
    \label{fig:exemplars}
\end{figure}

Inspired by the work in \cite{prabhu2019fonts}, we first generate an exemplar \textit{seed} dataset for each numeral system (Figure \ref{fig:exemplars}) from the corresponding Unicode characters. Following \cite{qi2006fuzzy}, from a category theoretic perspective these would constitute the classwise prototypes as they
\begin{enumerate}
\item Reflect the central tendency of the instances' properties or patterns; 
\item Are more similar to some category members than others; 
\item  Are themselves self-realizable but are not necessarily an instance. 
\end{enumerate}

In order to generate synthetic examples from these prototypical glyphs, we apply elastic deformations and corruptions (similar to \cite{mnistc} and as mentioned in \cite{simard2003best}) to the exemplars. We empirically chose elastic deformation parameters $\alpha = 8, \gamma \in [2, 2.5]$ in order to maximize variance while still retaining visual distinctness. The impact of these parameters on the synthetic image is explored in the Appendix (Figure \ref{fig:elasticGamma}).

\begin{figure}[h]
    \hspace*{-1.25cm}
    \centering
    \includegraphics[width=\linewidth]{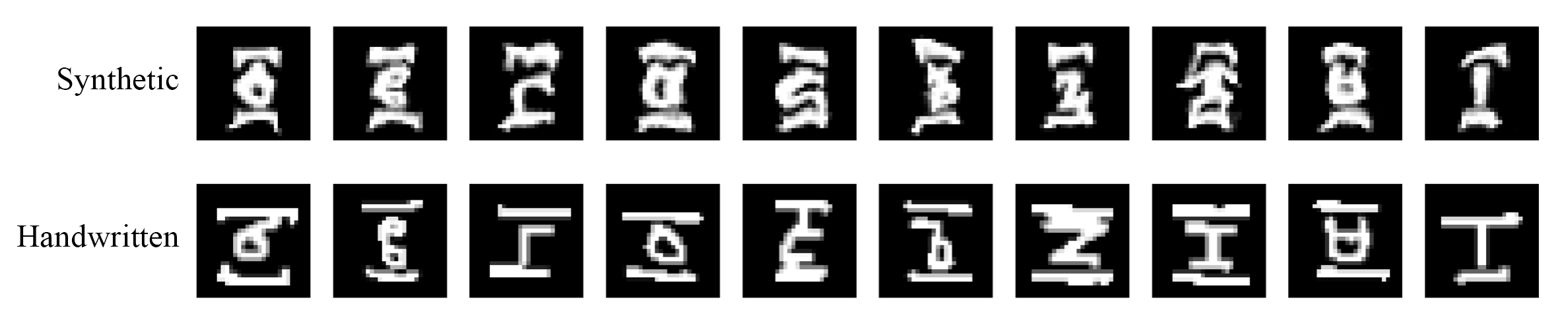}
    \caption{Comparison of Ge`ez-MNIST generated from exemplars and handwritten samples.}
    \label{fig:amhariccompare}
\end{figure}

We compared our results to a small dataset of written Ge`ez digits \citep{ethiopian-mnist}; the differences in exemplar versus handwritten data are shown in Figure \ref{fig:amhariccompare}. We note that, in cases where a limited amount of handwritten data is available, deformations and corruptions can be applied to those examples instead of Unicode exemplars.

\section{Dataset Details}
\begin{figure}[h]
    \hspace*{-1.25cm}
    \centering
    \includegraphics[width=\linewidth]{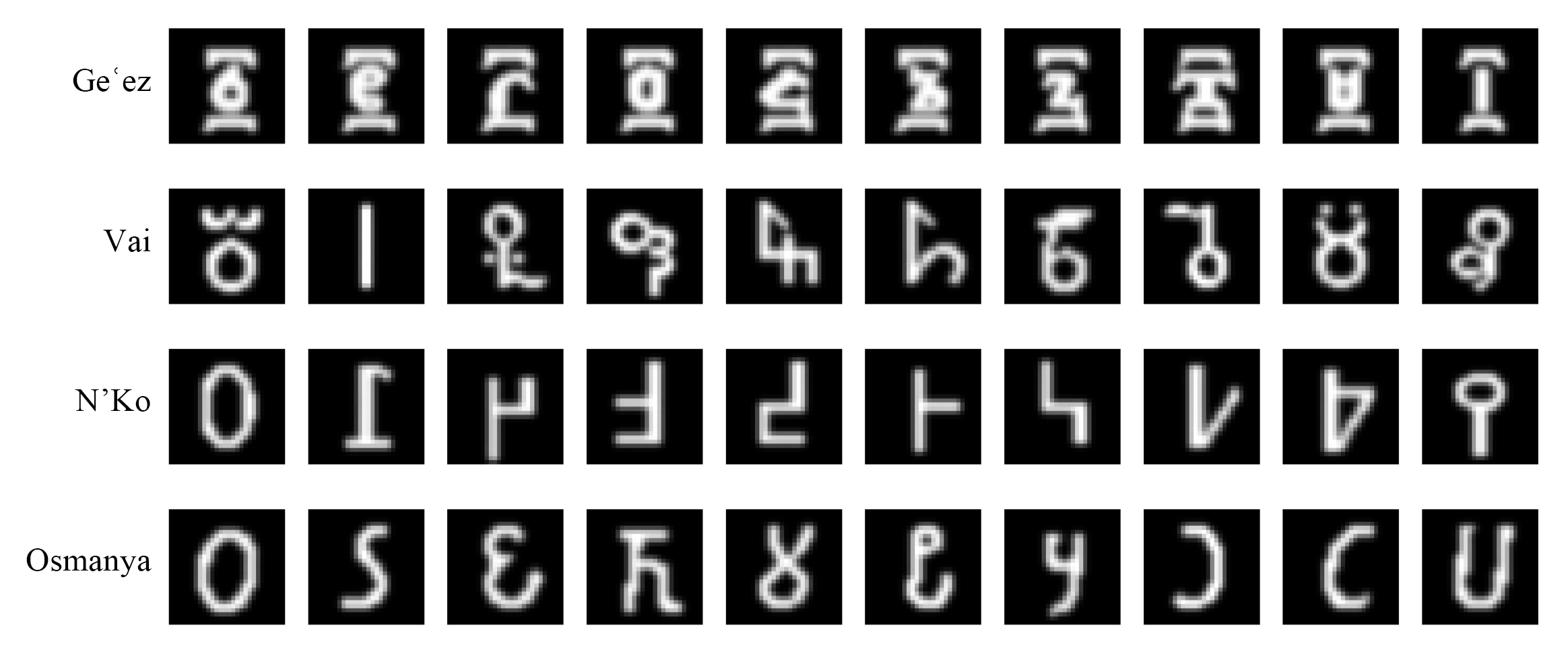}
    \caption{Global average of each glyph in Afro-MNIST datasets.}
    \label{fig:globalaverage}
\end{figure}

To produce ``drop-in'' replacements for MNIST, we closely emulate the format of the latter. Each of our datasets contains 60000 training images and 10000 testing images. Each image is greyscale and $28\times28$ pixels in size. Each dataset contains an equal number of images of each digit\footnote{The Ge`ez script lacks the digit 0, so our classes represent the numerals 1-10 for that script.}. To assess the robustness of our synthetic generation pipeline, we plot the global average of each character (Figure \ref{fig:globalaverage}) and note that it closely aligns with our exemplars. 

We are also interested in the morphological differences between our generated datasets and the original (Hindu-Arabic) MNIST dataset. We visualize the morphological characteristics of our datasets according to the methodology of \cite{castro2019morphomnist} and also plot the UMAP embeddings of the data. These analyses on Ge`ez-MNIST are shown in Figures \ref{fig:geezUMAP} and \ref{fig:geezMorphology}. Analogous analyses for our other datasets are given in the Appendix (Figures \ref{fig:vaiMorph} to \ref{fig:NKoMorph}).

\begin{figure}[h]
    \hspace*{-1.25cm}
    \centering
    \hspace*{1cm}                              
    \includegraphics[width=0.8\linewidth]{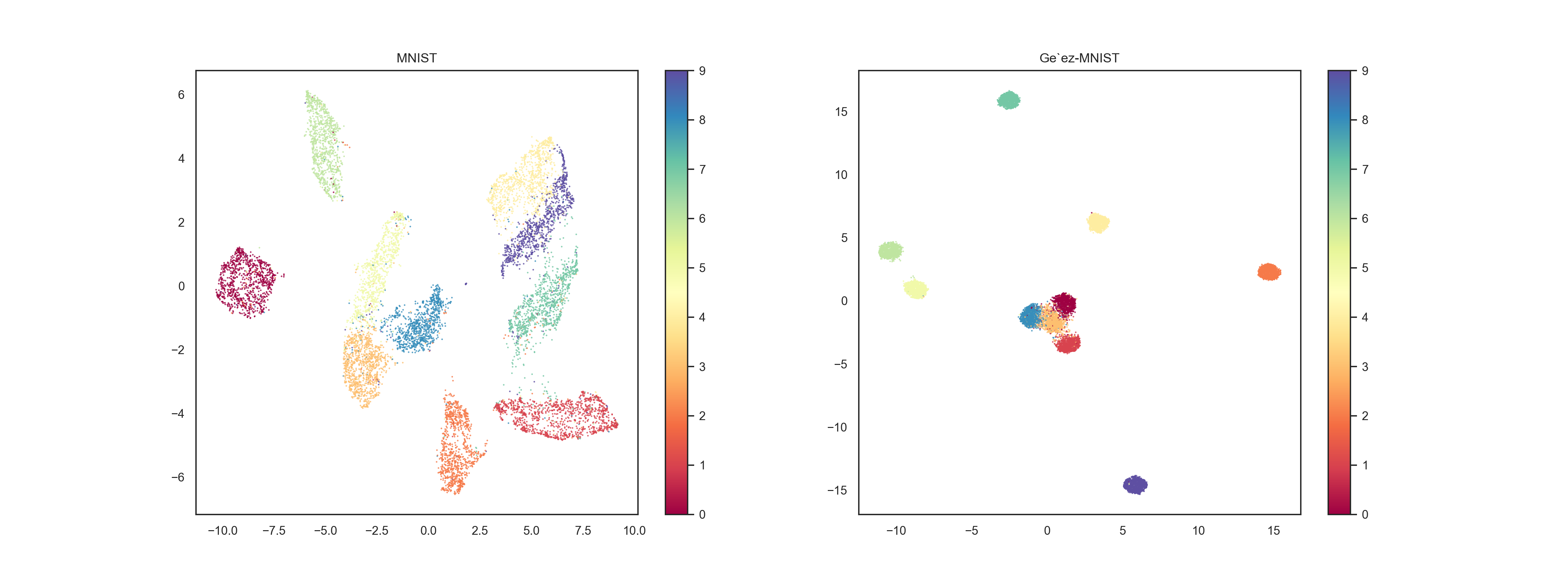}
    \caption{Comparison of UMAPs of MNIST and Ge`ez-MNIST.}
    \label{fig:geezUMAP}
\end{figure}

\begin{figure}[h]
    \centering
    \hspace*{-1cm}                              
    \includegraphics[width=0.7\linewidth]{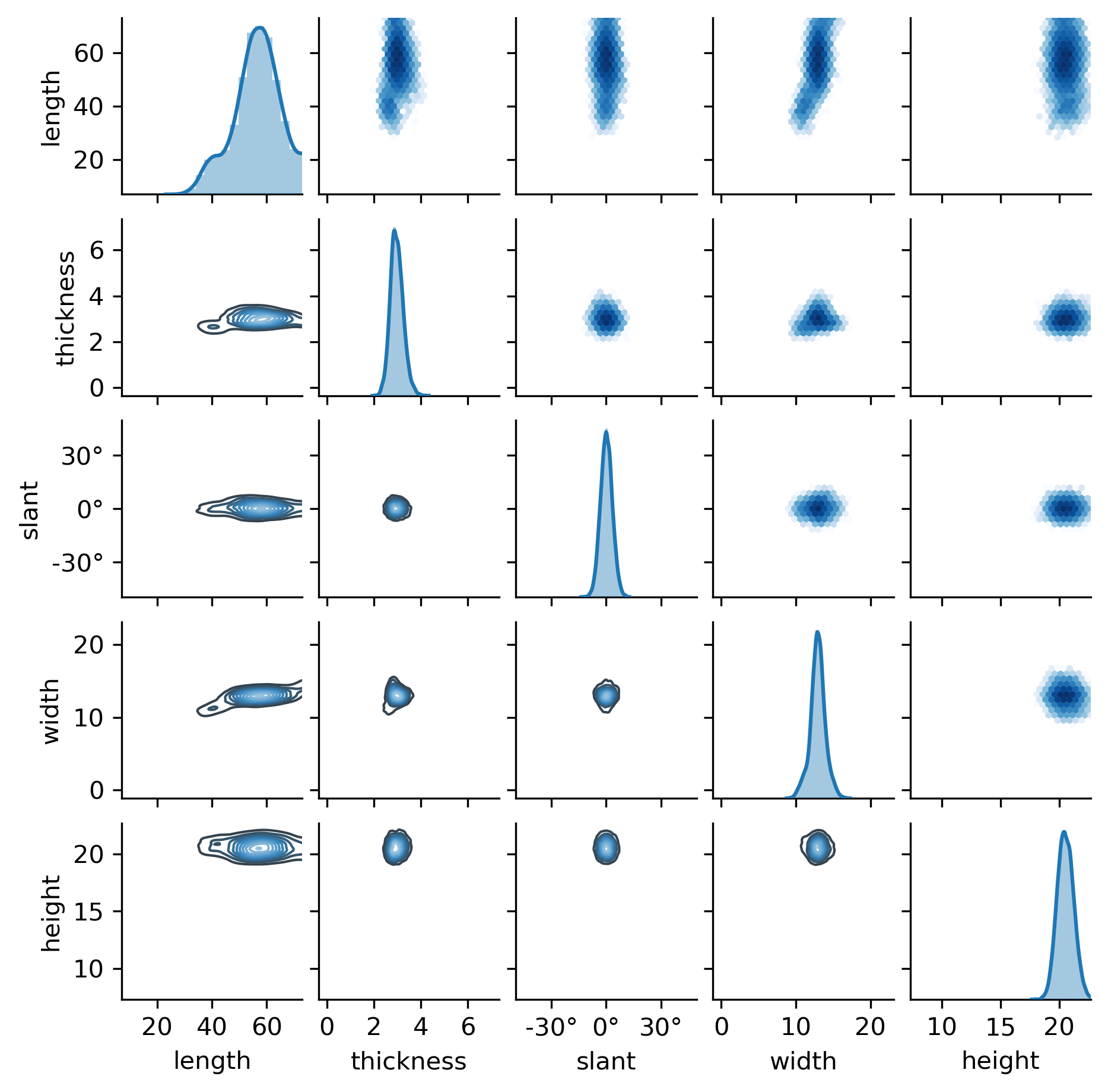}
\caption{Morphological comparison of MNIST versus Ge`ez-MNIST}
    \label{fig:geezMorphology}
\end{figure}

\section{Experiments}
To provide a baseline for machine learning methods on these datasets, we train LeNet-5 \citep{lecun1998gradient}, the network architecture first used on the original MNIST dataset, for numeral classification. We train with the Adam optimizer with an initial learning rate of $0.001$ using the categorical crossentropy loss. Models were trained until convergence. The model architecture is described in the Appendix (Table \ref{tab:LeNet}). Results are shown in Table \ref{tab:lenet}.

\begin{table}[h]
    \centering
    \begin{tabular}{|c|c|}
    \hline
        Dataset &  Accuracy (\%) \\
        \hline
        MNIST & 99.65\\
        Ge`ez-MNIST & 99.92\\
        Vai-MNIST & 100 \\
        Osmanya-MNIST & 99.99\\
        N'Ko-MNIST & 100 \\
        \hline
    \end{tabular}
    \caption{Accuracy of LeNet trained on our datasets}
    \label{tab:lenet}
\end{table}

\begin{figure}[H]
    \centering
    \includegraphics[width=\linewidth]{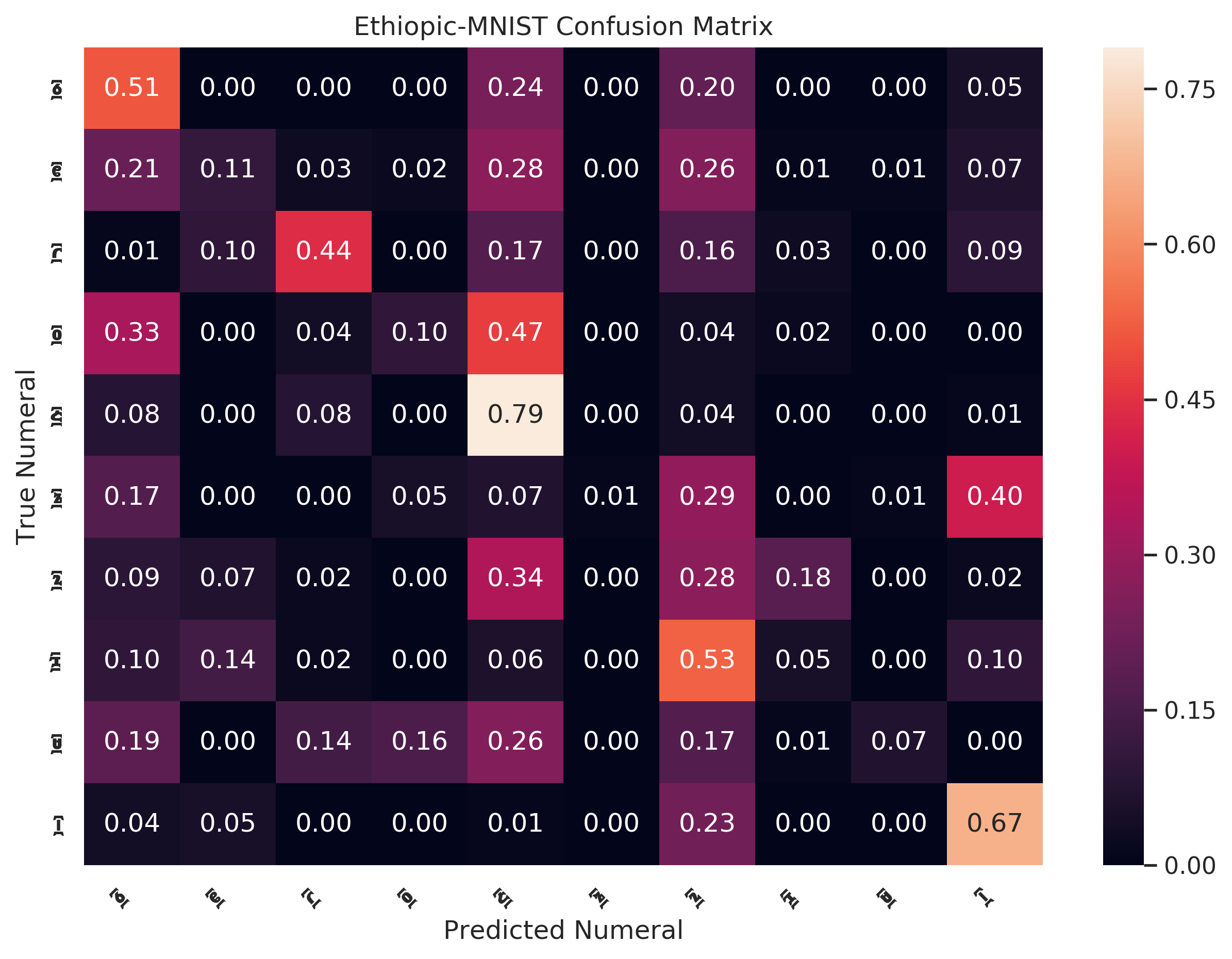}
    \caption{Confusion Matrix of LeNet-5 trained on the synthetic Ge`ez-MNIST, tested on the handwritten dataset.}
    \label{fig:cmats}
\end{figure}

We note that certain scripts have a comparatively high variability when written as opposed to our exemplars, and this leads to poor generalization (Figure \ref{fig:amhariccompare}, \ref{fig:geezMorphology}). Furthermore, it is clear from an examination of UMAP embeddings that our dataset is not as heterogenous as in-the-wild handwriting. After testing a LeNet-5 trained on Ge`ez-MNIST on the aforementioned dataset of handwritten Ge`ez numerals, we found the model achieved only an accuracy of 30.30\%. Certain numerals were easier to distinguish than others (Figure \ref{fig:cmats}). We expect this benchmark to be a fertile starting point for exploring augmentation and transfer learning strategies for low-resource languages.

\section{Conclusion}
We present Afro-MNIST --- a set of MNIST-style datasets for numerals in four African orthographies. It is our hope that the availability of these datasets enables the next generation of diverse research scientists to have their own ``Hello World" moment. We also open-source a simple pipeline to generate these datasets without any manual data collection, and we look forward to seeing the release of a wide range of new MNIST-style challenges from the machine learning community for other numeral systems.

\newpage

\bibliography{iclr2020_conference}

\begin{thebibliography}{11}
\providecommand{\natexlab}[1]{#1}
\providecommand{\url}[1]{\texttt{#1}}
\expandafter\ifx\csname urlstyle\endcsname\relax
  \providecommand{\doi}[1]{doi: #1}\else
  \providecommand{\doi}{doi: \begingroup \urlstyle{rm}\Url}\fi

\bibitem[Castro et~al.(2019)Castro, Tan, Kainz, Konukoglu, and
  Glocker]{castro2019morphomnist}
Daniel~C. Castro, Jeremy Tan, Bernhard Kainz, Ender Konukoglu, and Ben Glocker.
\newblock {Morpho-MNIST}: Quantitative assessment and diagnostics for
  representation learning.
\newblock \emph{Journal of Machine Learning Research}, 20, 2019.

\bibitem[Dosovitskiy et~al.(2015)Dosovitskiy, Fischer, Springenberg,
  Riedmiller, and Brox]{exemplarnets}
Alexey Dosovitskiy, Philipp Fischer, Jost~Tobias Springenberg, Martin
  Riedmiller, and Thomas Brox.
\newblock Discriminative unsupervised feature learning with exemplar
  convolutional neural networks.
\newblock \emph{IEEE transactions on pattern analysis and machine
  intelligence}, 38\penalty0 (9):\penalty0 1734--1747, 2015.

\bibitem[Eberhard et~al.(2019)Eberhard, Simons, and
  Fennig]{eberhard_simons_fennig_2019}
David~M. Eberhard, Gary~F Simons, and Charles~D Fennig.
\newblock Languages of the world, 2019.
\newblock URL \url{http://www.ethnologue.com/}.

\bibitem[LeCun et~al.(1998)LeCun, Bottou, Bengio, and
  Haffner]{lecun1998gradient}
Yann LeCun, L{\'e}on Bottou, Yoshua Bengio, and Patrick Haffner.
\newblock Gradient-based learning applied to document recognition.
\newblock \emph{Proceedings of the IEEE}, 86\penalty0 (11):\penalty0
  2278--2324, 1998.

\bibitem[Mgqwashu(2011)]{mgqwashu2011academic}
Emmanuel~Mfanafuthi Mgqwashu.
\newblock Academic literacy in the mother tongue: A pre-requisite for
  epistemological access.
\newblock \emph{Diversity, Transformation and Student Experience in Higher
  Education Teaching and Learning}, pp.\  159, 2011.

\bibitem[Molla(2019)]{ethiopian-mnist}
Tesfamichael Molla.
\newblock Ethiopian-mnist.
\newblock \url{https://github.com/Tesfamichael1074/Ethiopian-MNIST}, 2019.

\bibitem[Mu \& Gilmer(2019)Mu and Gilmer]{mnistc}
Norman Mu and Justin Gilmer.
\newblock Mnist-c: A robustness benchmark for computer vision.
\newblock \emph{arXiv preprint arXiv:1906.02337}, 2019.

\bibitem[Prabhu(2019)]{prabhu2019kannada}
Vinay~Uday Prabhu.
\newblock Kannada-mnist: A new handwritten digits dataset for the kannada
  language.
\newblock \emph{arXiv preprint arXiv:1908.01242}, 2019.

\bibitem[Prabhu et~al.(2019)Prabhu, Han, Yap, Douhaniaris, Seshadri, and
  Whaley]{prabhu2019fonts}
Vinay~Uday Prabhu, Sanghyun Han, Dian~Ang Yap, Mihail Douhaniaris, Preethi
  Seshadri, and John Whaley.
\newblock Fonts-2-handwriting: A seed-augment-train framework for universal
  digit classification.
\newblock \emph{arXiv preprint arXiv:1905.08633}, 2019.

\bibitem[Qi et~al.(2006)Qi, Zhu, Harrower, and Burt]{qi2006fuzzy}
Feng Qi, A-Xing Zhu, Mark Harrower, and James~E Burt.
\newblock Fuzzy soil mapping based on prototype category theory.
\newblock \emph{Geoderma}, 136\penalty0 (3-4):\penalty0 774--787, 2006.

\bibitem[Simard et~al.(2003)Simard, Steinkraus, Platt, et~al.]{simard2003best}
Patrice~Y Simard, David Steinkraus, John~C Platt, et~al.
\newblock Best practices for convolutional neural networks applied to visual
  document analysis.
\newblock In \emph{Icdar}, volume~3, 2003.

\end{thebibliography}
\bibliographystyle{iclr2020_conference}

\appendix
\section{Appendix}

\begin{figure}
    \centering
    \hspace{-1cm}   \includegraphics[width=0.8\linewidth]{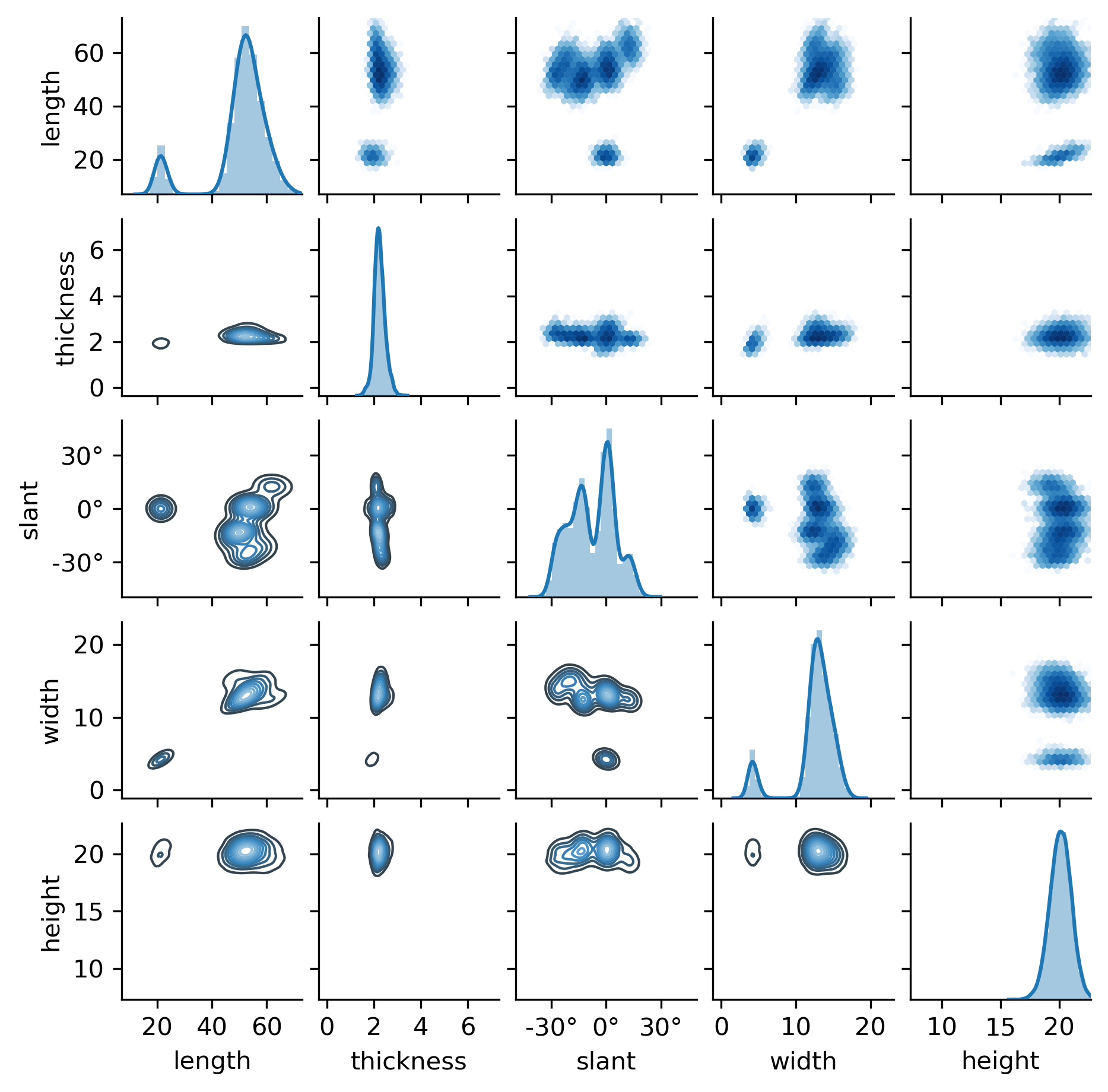} \\
    \hspace*{1cm}                      \includegraphics[width=0.9\linewidth]{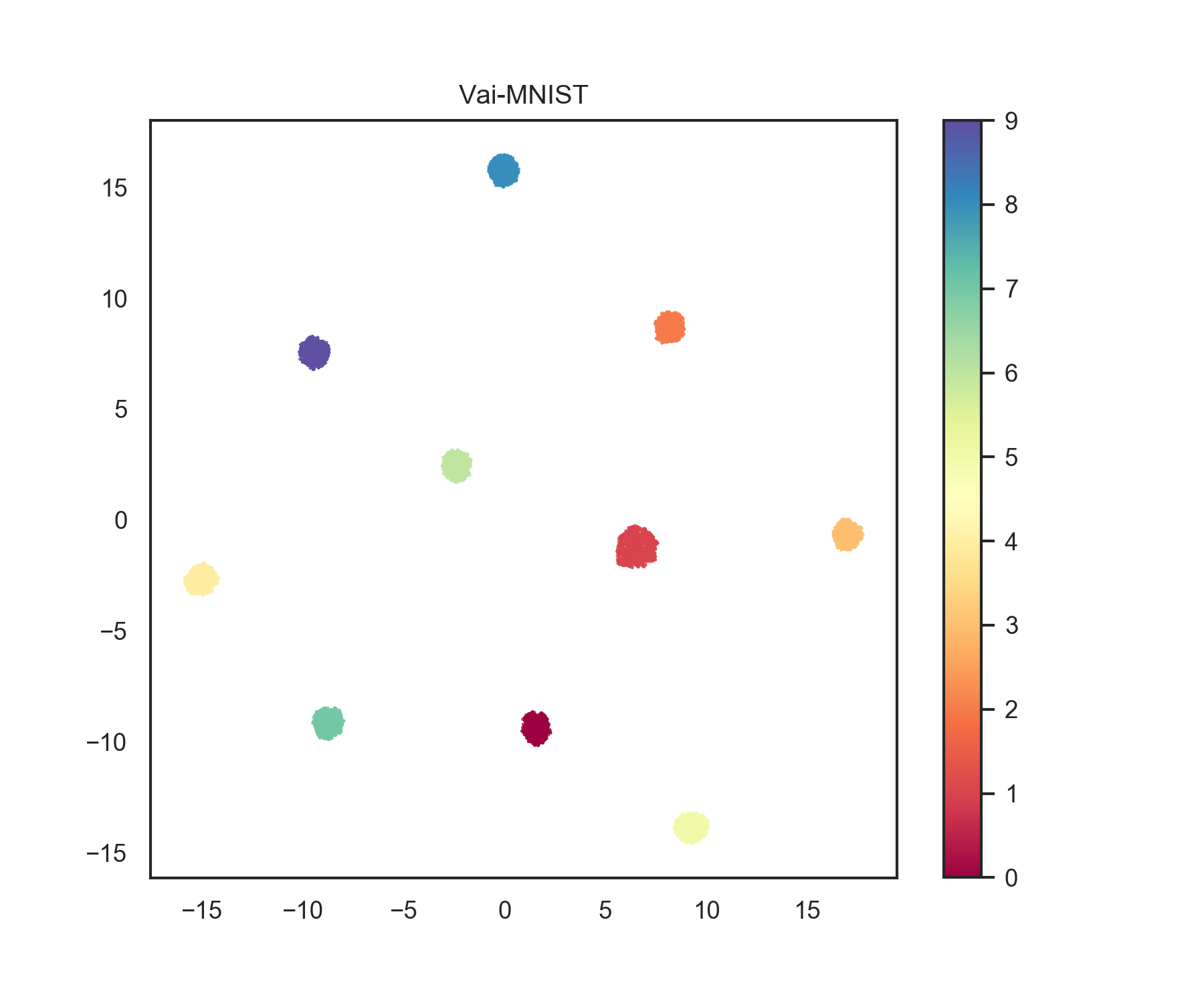}
    \caption{Vai-MNIST morphology and UMAP embedding.}
    \label{fig:vaiMorph}
\end{figure}

\begin{figure}
    \centering
    \hspace{-1cm}   \includegraphics[width=0.8\linewidth]{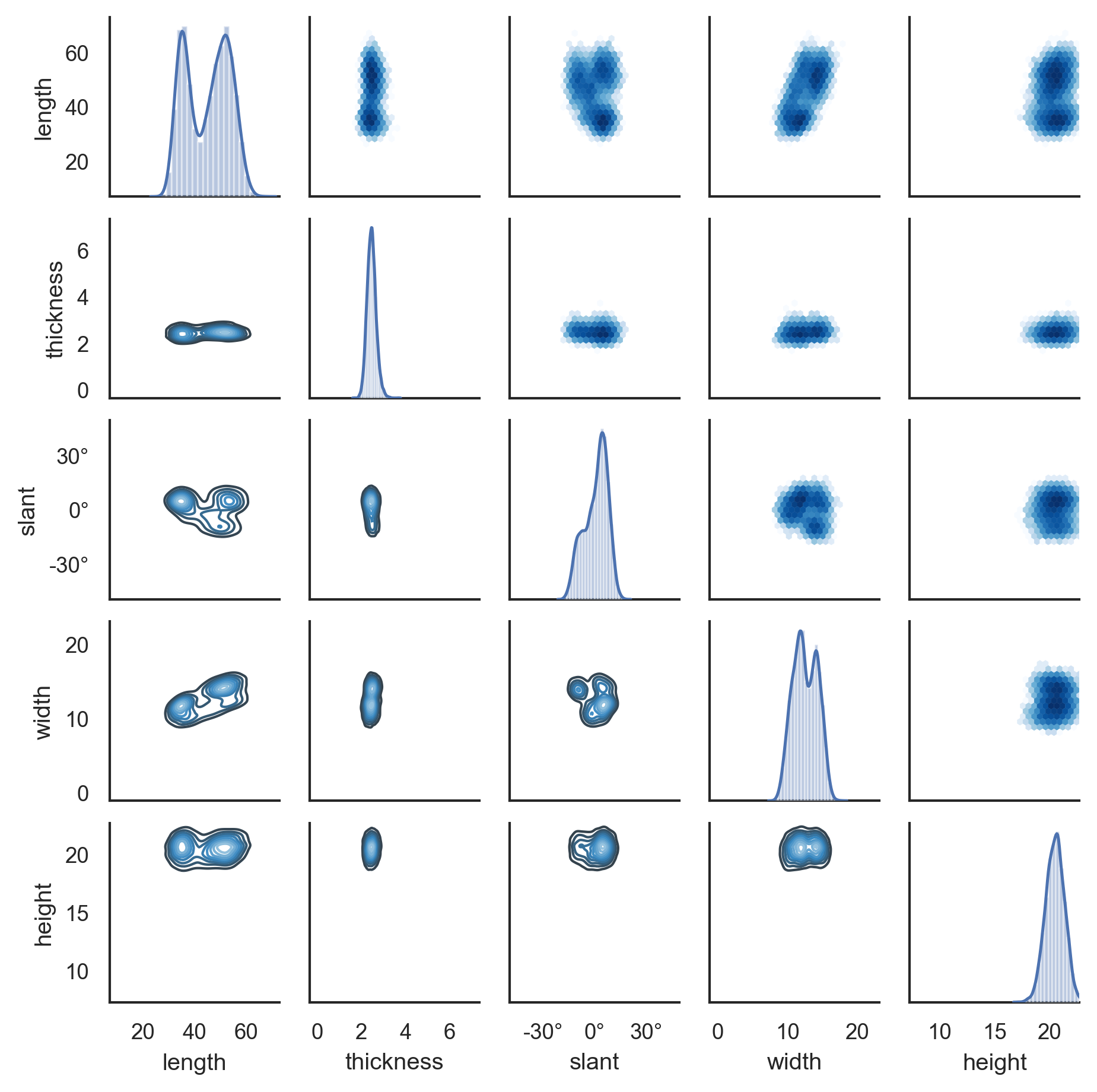} \\
    \hspace*{1cm}                      \includegraphics[width=0.9\linewidth]{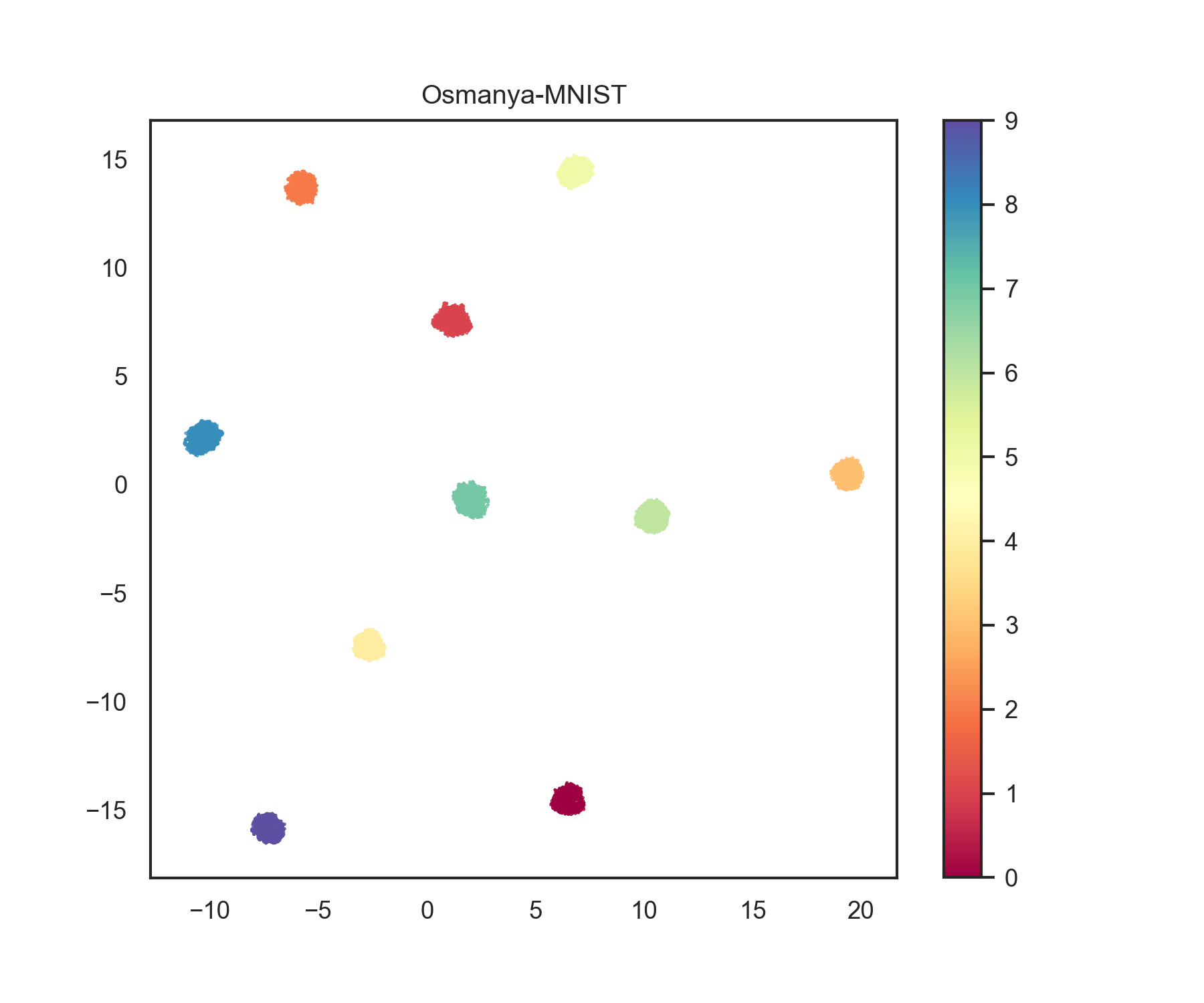}
    \caption{Osmanya-MNIST morphology and UMAP embedding.}
    \label{fig:OsmanyaMorph}
\end{figure}

\begin{figure}
    \centering
    \hspace{-1cm}   \includegraphics[width=0.8\linewidth]{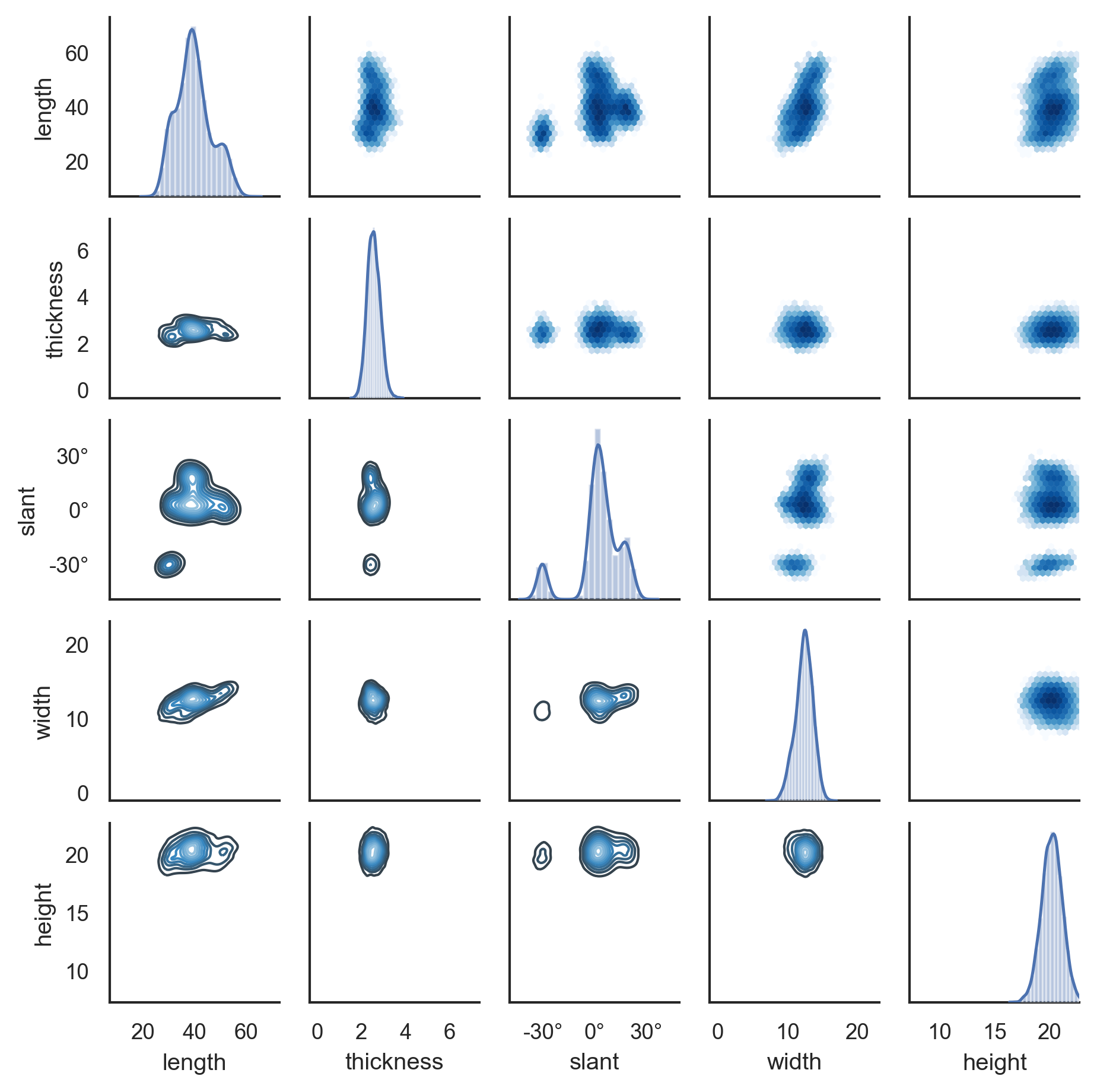} \\
    \hspace*{1cm}                      \includegraphics[width=0.9\linewidth]{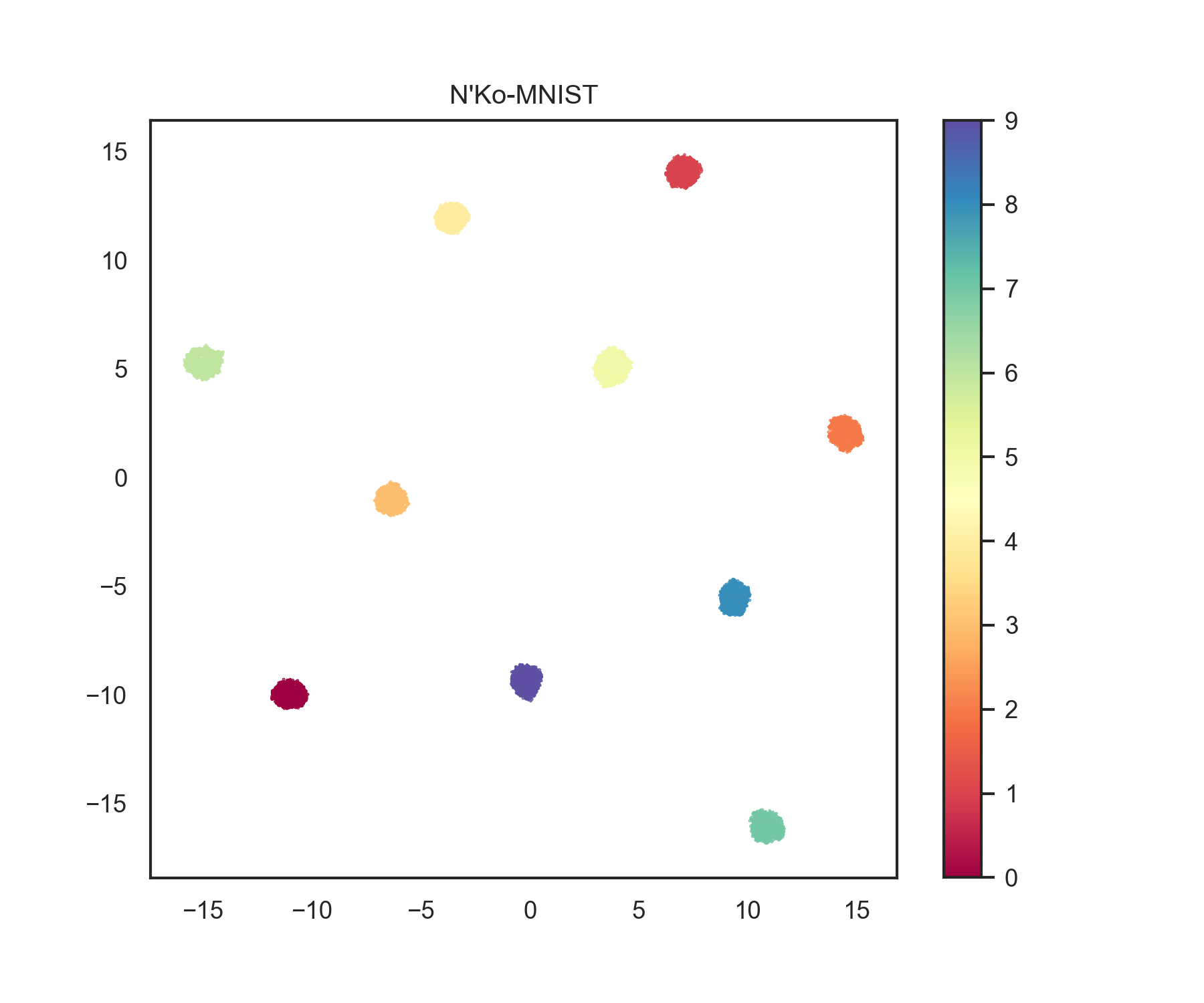}
    \caption{N'Ko-MNIST morphology and UMAP embedding.}
    \label{fig:NKoMorph}
\end{figure}

\begin{figure}[h]
    \centering
    \includegraphics[width=\linewidth]{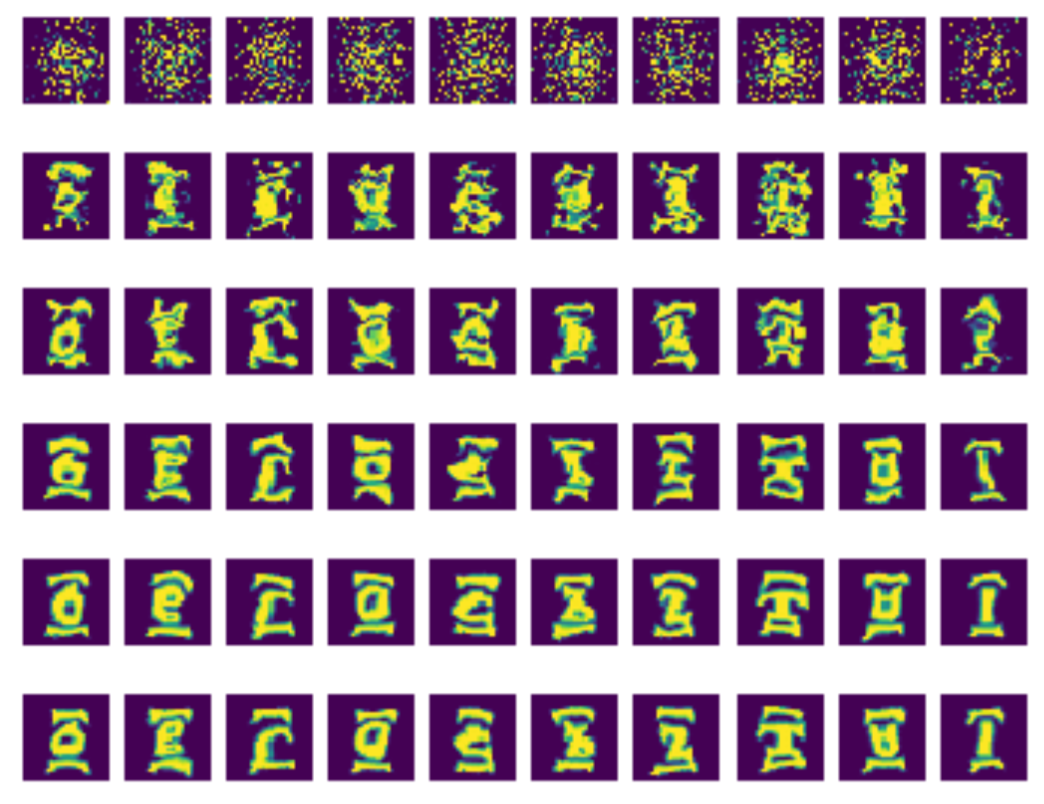}
    \caption{Impact of $\gamma$ on generated Ge`ez numerals. From top to bottom, $\gamma = 0.01, 1, 1.5, 2, 2.5, 3$}
    \label{fig:elasticGamma}
\end{figure}

\begin{table}[h]
    \centering
    \begin{tabular}{|c|c|c|c|}
    \hline
        Layer & Filter Size & Units & Parameters \\
    \hline
        Conv2D & 5x5 & 6 & 156 \\
        Max Pool & 2x2 & 6 & 0 \\
        Tanh & & & 0\\
        Conv2D & 5x5 & 16 & 2416\\
        Max Pool & 2x2 & 6 & 0 \\
        Tanh & & 0 & \\
        Flatten & & &\\
        Dense &  & 120 & 30804\\
        Tanh & & & 0\\
        Dense &  & 84 & 10164\\
        Tanh & & & 0 \\
        Dense & & 10 & 850\\
        Softmax & & & \\
    \hline
    \end{tabular}
    \caption{The architecture of LeNet from \cite{lecun1998gradient}}
    \label{tab:LeNet}
\end{table}
\end{document}